\documentclass[letterpaper,journal]{article} 
\usepackage[submission]{aaai24}  
\usepackage{times}  
\usepackage{helvet}  
\usepackage{courier}  
\usepackage[hyphens]{url}  
\usepackage{graphicx} 
\usepackage{natbib}  
\usepackage{caption} 
\usepackage{algorithm}
\usepackage{algorithmic}

\usepackage{newfloat}
\usepackage{listings}
\DeclareCaptionStyle{ruled}{labelfont=normalfont,labelsep=colon,strut=off} 
\floatstyle{ruled}
\newfloat{listing}{tb}{lst}{}
\floatname{listing}{Listing}
\usepackage{makecell}
\usepackage{multirow}
\usepackage{array}
\usepackage{amsmath,amssymb,amsfonts}
\usepackage{bbding}
\usepackage{url}
\usepackage{makecell, multirow, tabularx}
\usepackage{pifont}
\usepackage{makecell}

\title{SAMCT: Segment Any CT Allowing Labor-Free Task-Indicator Prompts}
\author{
	Xian Lin\textsuperscript{\rm 1}, Yangyang Xiang\textsuperscript{\rm 1}, Zhehao Wang\textsuperscript{\rm 1}, Kwang-Ting Cheng\textsuperscript{\rm 2}, Zengqiang Yan\textsuperscript{\rm 1*}, Li Yu\textsuperscript{\rm 1}
}
\affiliations{
	\textsuperscript{\rm 1}School of Electronic Information and Communications, Huazhong University of Science and Technology,\\
	\textsuperscript{\rm 2}School of Engineering, Hong Kong University of Science and Technology
	
}

\usepackage{bibentry}

\begin{document}
	
	\maketitle
	
	\begin{abstract}
		Segment anything model (SAM), a foundation model with superior versatility and generalization across diverse segmentation tasks, has attracted widespread attention in medical imaging. However, it has been proved that SAM would encounter severe performance degradation due to the lack of medical knowledge in training and local feature encoding. Though several SAM-based models have been proposed for tuning SAM in medical imaging, they still suffer from insufficient feature extraction and highly rely on high-quality prompts. In this paper, we construct a large CT dataset consisting of 1.1M CT images and 5M masks from public datasets and propose a powerful foundation model SAMCT allowing labor-free prompts. Specifically, based on SAM, SAMCT is further equipped with a U-shaped CNN image encoder, a cross-branch interaction module, and a task-indicator prompt encoder. The U-shaped CNN image encoder works in parallel with the ViT image encoder in SAM to supplement local features. Cross-branch interaction enhances the feature expression capability of the CNN image encoder and the ViT image encoder by exchanging global perception and local features from one to the other. The task-indicator prompt encoder is a plug-and-play component to effortlessly encode task-related indicators into prompt embeddings. In this way, SAMCT can work in an automatic manner in addition to the semi-automatic interactive strategy in SAM. Extensive experiments demonstrate the superiority of SAMCT against the state-of-the-art task-specific and SAM-based medical foundation models on various tasks. The code, data, and models will be released at \url{https://github.com/xianlin7/SAMCT}.
	\end{abstract}
	
	\section{Introduction}
	\label{sec:introduction}
	
	Computed tomography (CT), a medical imaging modality that provides detailed visualization of the internal anatomical structures within the human body, plays a pivotal role in the clinical diagnosis of diseases~\cite{ctimage}. In contrast to alternative medical imaging modalities, CT has a faster scanning speed and is widely used to examine different body structures, including head, chest, abdomen, and limbs~\cite{samed2d}. Through the segmentation of anatomical structures or pathological lesions within CT images, medical practitioners are empowered to precisely discern and pinpoint regions of interest, thereby enhancing the accuracy of diagnoses and making more efficacious treatment strategies. However, pixel-wise segmentation is expertise-dependent, time-consuming, and labor-intensive. Recently, though deep learning-based models, represented by convolutional neural networks (CNN) and vision transformer (ViT), have exhibited remarkable potential in automating CT image segmentation, such models are typically tailored and trained for specific tasks, excelling in the designated tasks but struggling to generalize effectively across a spectrum of different tasks~\cite{unet,nnunet,canet}. The inherent task-specific nature hinders their widespread deployment in the realm of CT image analysis, where diverse tasks are expected. 
	
	\begin{table*}[!t]
		\centering
		\caption{Comparison of foundation models designed for universal medical image segmentation. Our SAMCT is a comprehensive model that supplements local feature encoding and allows labor-free prompts}\label{tab1}
		\resizebox{\textwidth}{!}{
		\begin{tabular}{c|ccccccc}
			\hline
			Method   & Input size & Prompts        & Labor-free mode & ViT encoder    & CNN encoder & Prompt encoder & Mask decoder   \\ \hline
			MedSAM~\cite{medsam}       & 1024×1024  & box            & \ding{55}           & fine-tuning    & \ding{55}       & frozen         & fine-tuning    \\
			SAMed~\cite{samed}        & 512×512    & -           & -           & LoRA tuning    & \ding{55}        & fine-tuning    & fine-tuning    \\
			MSA~\cite{msa}          & 1024×1024  & point          & \ding{55}           & adapter tuning & \ding{55}        & frozen         & adapter tuning \\
			SAM-Med2D~\cite{samed2d}    & 256×256    & point\&box\&mask & \ding{55}          & adapter tuning & \ding{55}        & fine-tuning    & fine-tuning    \\
			\textbf{SAMCT}        & 256×256    & point/box/task-indicator & \Checkmark          & adapter tuning & \Checkmark       & frozen         & forzen         \\ \hline
		\end{tabular}
	}
	\end{table*}
	
	The foundation model, a sophisticated neural network trained on extensive datasets, possesses the capacity to capture the fundamental attributes of data and abstract them at a high level~\cite{foundationmodel}. This extraordinary representation ability equips foundation models with exceptional versatility and competence to generalize effectively, facilitating the amalgamation of diverse segmentation tasks into a singular model~\cite{samed2d}. The Segment Anything Model (SAM) stands as the most renowned foundational model specifically crafted for segmentation tasks~\cite{sam}, attracting numerous studies aimed at assessing the performance of SAM on medical images~\cite{evalsam2}. However, these works have revealed the apparent inferior performance of SAM in medical images compared to natural scenes, particularly when dealing with targets characterized by low contrast, faint boundaries, intricate geometries, and diminutive dimensions~\cite{evalsam1}. It is due to the following three reasons: the absence of medical image data in the training phase, the lack of semantic understanding, and the loss of local contextual information due to its large tokenization size~\cite{evalsam3}. To adapt SAM to medical scenarios, some works apply visual tuning techniques to SAM on downstream medical datasets~\cite{medsam,samed,msa} but may make SAM lose its great versatility due to the singularity of training data~\cite{polypsam}. Besides, SAM-based methods require accurate point/box/mask prompts to generate accurate segmentation, resulting in a semi-automatic manner, and still fails to escape from extensive labor and expertise~\cite{sam,evalsam1}.
	
	In this paper, we propose SAMCT, a foundation model supplementing local features to SAM to segment any CT and allowing both manual prompt mode and labor-free prompt mode. Specifically, SAMCT consists of the original SAM, a U-shaped CNN module, and a cross-branch interaction module. To inherit the strong feature representation ability of SAM, the structure and parameters of SAM are kept and frozen. To better recognize the objects with complex shapes, low contrast, and various sizes in CT images, we introduce a U-shaped CNN sub-network to capture local features, which is placed in parallel with the ViT image encoder of SAM. Then, the cross-branch interaction module is strategically positioned between the ViT branch and the U-shaped CNN branch, facilitating the augmentation of their feature representation capacities by exchanging global semantics and local information from one to the other. Furthermore, we extend SAMCT into an automated segmentation model by introducing an optional lightweight task-indicator prompt encoder. It serves as a plug-and-play module, allowing SAMCT to identify the object for segmentation under the prompt of only task indicators associated with target objects (\textit{e.g.}, COVID-19, spleen, liver, \textit{etc}). In this way, SAMCT can accurately identify various objects even without manual prompts. More importantly, to comprehensively testify SAMCT and benefit future research, we construct a substantial CT dataset named CT5M, comprising 1.1M images and 5M masks from 30 publicly accessible datasets and including 118 objects covering head\&neck, chest, abdomen, pelvis, lesion, bone, and vessel. Experimental results demonstrate the superiority of SAMCT against the state-of-the-art (SOTA) foundation models and task-specific models in both versatility and generalization.
	
	Our contributions can be summarized as:
	
	\begin{itemize}
		
		\item We propose a modality-specific foundational model, SAMCT, tailored for universal segmentation within the CT modality, achieving superior performance on intricate anatomical structures and ambiguous lesions.
		
		\item We design a U-shaped CNN branch and a cross-branch interaction module to augment the feature representation capacity of the image encoder in SAM.
		
		\item We propose a labor-free and plug-and-play prompt module, namely the task-indicator prompt encoder, to facilitate pixel-wise segmentation even without labor-intensive and precise prompts, which is particularly valuable in clinical applications.
		
		\item We construct and release a large CT dataset by collecting, pre-processing, and standardizing 30 public datasets, consisting of 5821 patients, 1.1M images, and 5M image-mask pairs, and covering 118 objects.
		
	\end{itemize}

	\section{Related Work}
	\label{sec:relatedwork}
	
	\subsection{Medical Image Segmentation}
	
	Deep learning-based methods have been widely explored in medical image segmentation~\cite{segmentationsurvey2021} which can be roughly categorized into CNN-based~\cite{aaunet}, Transformer-based~\cite{missformer}, and CNN-Transformer hybrid~\cite{transunet}. Among CNN-based models, U-Net is the seminal work due to its prominent ability to preserve fine details~\cite{unet}. Inspired by U-Net, a series of improvements have been proposed, including introducing residual blocks and attention blocks~\cite{aaunet}. With the great success of transformer on computer vision tasks, transformer-based/-hybrid models have been introduced to medical image segmentation. TransUNet has been one of the most representative methods by introducing transformer layers at the bottom of the U-Net encoder~\cite{transunet}. A common theme across these models is to construct a U-shaped framework with skip connections to preserve important local features.
	
	\subsection{SAMs for Medical Imaging}
	
	The impressive versatility and generalization of SAM have caused revolutionary changes in the field of medical image segmentation~\cite{evalsam1}. The promptable segmentation paradigm and large training data make it possible to address multiple medical segmentation tasks by a unified model~\cite{samed2d}. A series of studies have been conducted to evaluate the performance of SAM in medicine~\cite{evalsam2}. Compared with natural scenes, the performance of SAM in medical scenes deteriorates sharply due to the lack of medical training data and sufficient local features~\cite{evalsam3}. Mainstream vision tuning methods have been utilized to adapt SAM to medicine, including fine-tuning different components~\cite{medsam,polypsam}, introducing adapters~\cite{msa}, and applying low-rank-based (LoRA) strategies~\cite{samed}. Such methods focus on transferring SAM from natural to medical scenes with limited learnable parameters or affordable structure adjustments. Major differences between SAMCT and SAM-based methods for universal medical image segmentation are summarized in Table~\ref{tab1}. In general, containing richer local features (\textit{i.e.}, details) and supporting labor-free prompts make SAMCT special. 
	
	\begin{figure*}[ht]
		\centering{\includegraphics[width=1\textwidth]{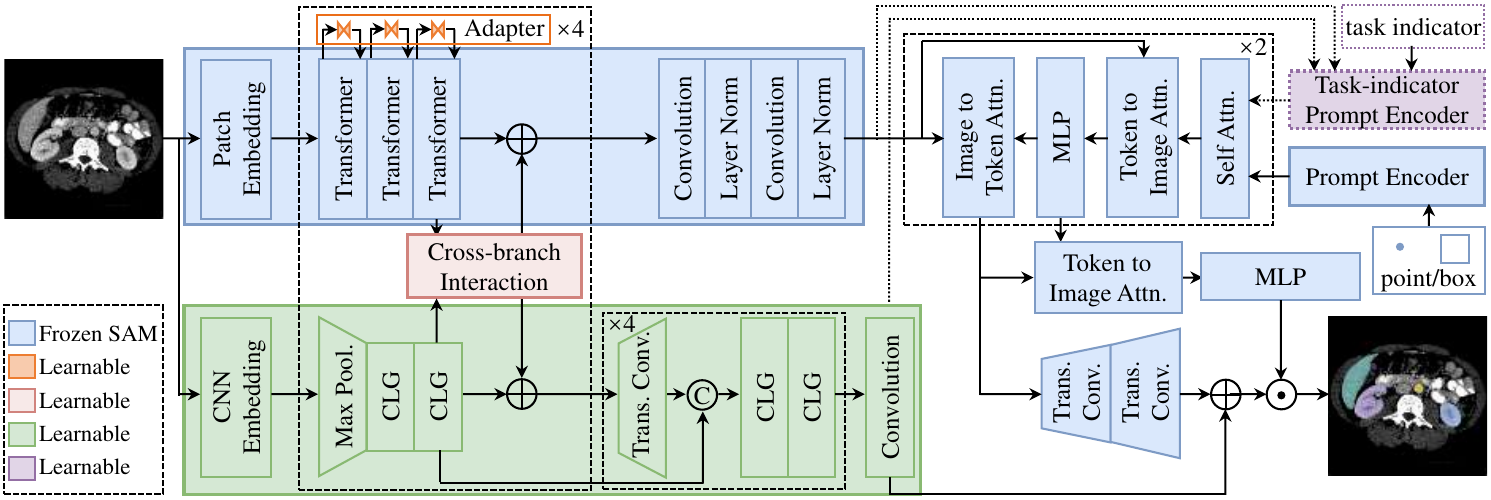}}
		\caption{
			Overview of SAMCT. Modules painted in green, light red, and purple are the U-shaped CNN image encoder, the cross-branch interaction module, and the task-indicator prompt encoder respectively. All modules from SAM are painted in blue and frozen during training. CLG represents the combination of convolution, layer normalization, and Gelu. Trans.Conv. represents transpose convolution.
		}
		\label{fig1}
	\end{figure*} 
	
	\subsection{Prompts in SAMs} 
	
	Accurate segmentation of SAMs relies on precise manual prompts, \textit{i.e.}, points, bounding boxes, or masks, which brings great inconvenience to its clinical deployment~\cite{evalsam1}. There are two ways to reduce the workload of SAMs. One way is to redesign the prompt encoders to generate prompt embeddings without human intervention. Auto-prompting SAM generates prompt embeddings from the image embedding by a down-up FCN subnetwork~\cite{autoprompt}. SurgicalSAM develops a prototype-based prompt encoder to generate dense and sparse prompt embeddings~\cite{prototypeprompt}. Despite their effectiveness, such prompt encoders are deeply coupled to the mask decoder, undermining the universality of SAM. The other way is to introduce additional object detectors or segmentation networks to obtain point or bounding box prompts. TongueSAM proposes a prompt generator based on object detection~\cite{tonguesam}. UV-SAM produces mask and box prompts from the outputs of Segformer~\cite{uvsam}. Similarly, AutoSAM uses a Harmonic Dense Net to generate prompt embeddings from input images~\cite{autosam}. Though these methods can protect the universality of the mask decoder, they need an additional heavy network to generate corresponding prompts for each downstream task, making the foundation model degenerate into a heavy ensemble model. Comparatively, SAMCT is equipped with an optional and lightweight task-indicator prompt encoder, which works as a plug-and-play component and is only related to target objects and independent of SAMCT itself.
	
	\section{Method}
	\label{sec:method}

	\subsection{Overview}
	
	As depicted in Fig.~\ref{fig1}, SAMCT is composed of the original SAM, a U-shaped CNN image encoder, a cross-branch interaction module, and an optional task-indicator prompt encoder. All parameters of SAM are frozen during training. To align the encoded feature space in SAM with medical imaging, we parallelize a simple adapter on the feed-forward layer of each transformer in the ViT image encoder of SAM. The adapter consists of a linear projection with $4\times$ channel decrease, a Gelu activation function, and a linear projection with $4\times$ channel increase. The U-shaped CNN image encoder, a branch placed in parallel with the ViT image encoder of SAM, is introduced to capture local features in high-resolution feature maps. Between them, the cross-branch interaction module is developed to transfer local information and global perception from one to the other. The encoded feature embeddings, together with the encoded prompt embeddings from the prompt encoder of SAM, flow into the mask decoder for final segmentation. In terms of the mask decoder, we fuse the full-resolution image features from CNN and the up-sampled ViT image embeddings before computing the final mask by dot product for further local feature enhancement. Combining with SAM, the U-shaped CNN image encoder and the cross-branch interaction module adapt SAM for segmenting any CT. Furthermore, to ease the usage of SAMCT in clinical scenarios, we propose a plug-and-play task-indicator prompt encoder to replace the manual prompt encoder and automatically produce effective prompt embeddings for each specific task.
	
	\subsection{U-shaped CNN Image Encoder}
	
	The design of the CNN image encoder is inspired by U-Net \cite{unet}, which is proven to be more consistent with the characteristics of medical imaging as illustrated in Fig.~\ref{fig1}.
	Given any input $x \in \mathbb{R} ^ {3 \times H \times W}$, the HAAM module proposed by AAU-net~\cite{aaunet} is first used to extract its CNN embedding $F_c \in \mathbb{R} ^ {d_c \times H \times W}$. Then, four encoding blocks are executed in sequence to continuously adjust its resolution and obtain deeper semantic features. Each encoding block consists of a max pooling, two sequential convolutions with the stride of 1 and the kernel size of 3$\times$3, a layer normalization, and a Gelu activation. Besides, the previous convolution doubles the input channels while the second convolution maintains its input channels. After such encoding blocks, we obtain the smallest feature map $F_{16} \in \mathbb{R} ^ {16d_c \times \frac{H}{16} \times \frac{W}{16}}$ of the CNN-branch. Then, four decoding blocks are executed sequentially to recover the resolution and capture important local features. Each decoding block starts with a transpose convolution with half the number of channels, a stride of 2, and a kernel size of 2. Then, the output of transpose convolution is concatenated with the same-resolution feature map from the encoding block through skip connection. After concatenation, two sequential convolutions, a layer normalization, and a Gelu activation are applied. During decoding, another four feature maps with different resolutions $F_{32} \in \mathbb{R} ^ {8d_c \times \frac{H}{8} \times \frac{W}{8}}, F_{64} \in \mathbb{R} ^ {4d_c \times \frac{H}{4} \times \frac{W}{4}}, F_{128} \in \mathbb{R} ^ {2d_c \times \frac{H}{2} \times \frac{W}{2}}, F_{256} \in \mathbb{R} ^ {d_c \times H \times W}$ are generated. Finally, a $1 \times 1$ convolution is to adjust the channel number to 32 to match that of the full-resolution image embedding in the mask decoder.
	
	\subsection{Cross-branch Interaction}
	
	\begin{figure}[!t]
		\centering{\includegraphics[width=\linewidth]{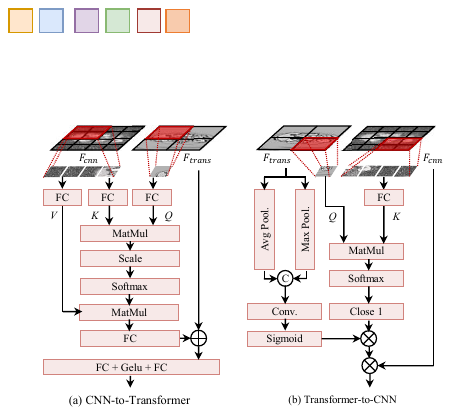}}
		\caption{
			Two flows of the cross-branch interaction module.
		}
		\label{fig2}
	\end{figure}
	
	Cross-branch interaction module builds two information flows between the ViT image encoder and the CNN image encoder, including the CNN-to-Transformer flow and the Transformer-to-CNN flow as depicted in Fig.~\ref{fig2}. The inputs of cross-branch interaction include an image embedding $F_{cnn}$ from a certain scale of the CNN branch and an image embedding $F_{trans}$ from the ViT branch correspondingly. After aligning $F_{cnn}$ and $F_{trans}$, each patch $F_p \in \mathbb{R} ^ {1 \times d_p}$ in $F_{trans}$ correspond to a $k \times k$ window $F_w \in \mathbb{R} ^ {k^2 \times d_w}$ in $F_{cnn}$ as shown in the red regions of Fig.~\ref{fig2}. In the CNN-to-Transformer flow, through three different projections, each patch $F_{p}$ can be recognized as a query (\textit{i.e.}, $Q$) and the corresponding $k \times k$ region $F_w$ can be recognized as key (\textit{i.e.}, $K$) and value (\textit{i.e.}, $V$) to build dependency via cross-attention, formulated as:
	\begin{equation}
		F_c = \text{\textbf{Softmax}}(\frac{F_pE_q(F_wE_k)^T}{\sqrt{d_p}})(F_wE_v),
		\label{eq1}
	\end{equation}
	where $E_k, E_v \in \mathbb{R} ^ {d_w \times d_p}$, and $E_q \in \mathbb{R} ^ {d_p \times d_p}$ are projection matrices. By combining $F_p$ with $F_{c}$ after projection, detailed/local features can flow from the CNN branch to the ViT branch. 
	
	In the Transformer-to-CNN flow, $F_{trans}$ can remind $F_{cnn}$ of regions demanding more attention across the entire image space based on its superior global perception. Therefore, we generate a spatial attention map to augment $F_{cnn}$ by a coarse-to-fine attention mechanism. First, we apply max pooling and average pooling onto $F_{trans}$ to generate a rough spatial attention map, where each pixel corresponds to a $k \times k$ window of $F_{cnn}$. To refine the attention in each $k \times k$ window of $F_{cnn}$, we compute the correlation between the patch $F_p$ and the corresponding window $F_w$ and normalize the value to (0.5, 1.5), formulated as:
	\begin{equation}
		\begin{aligned}
			A_w =\text{\textbf{Softmax}}(\frac{F_p(F_wE_w)^T}{\sqrt{d_p}}), \\
			A^*_w = \frac{e^{A_w - k^{-2}}}{1 + e^{A_w - k^{-2}}} + 0.5,
		\end{aligned}
		\label{eq2}
	\end{equation}
	where $E_w  \in \mathbb{R} ^ {d_w \times d_p}$ is a projection matrix to unify the dimensions of $F_p$ and $F_w$. After broadcasting the coarse attention $A_p \in \mathbb{R} ^ {1 \times 1}$ of $F_p$ into $A^*_p \in \mathbb{R} ^ {k \times k}$, $A^*_w$ is multiplied by $A^*_p$ to obtain the refined attention map of $F_w$. Finally, multiplying the entire refined attention map with $F_{cnn}$ can supplement global information from the ViT branch to the CNN branch.
	
	\subsection{Task-indicator Prompt Encoder}
	
	\begin{figure}[!t]
		\centering{\includegraphics[width=1\linewidth]{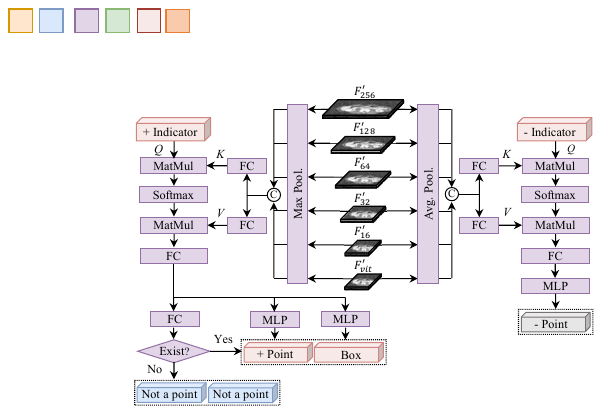}}
		\caption{
			Details of the task-indicator prompt encoder.
		}
		\label{fig3}
	\end{figure}
	
	\begin{table*}[!t]
		\centering
		\caption{Composition of CT5M. Datasets with $^*$  are invisible during training and are only used for generalization verification.}
		\resizebox{\textwidth}{!}{
		\begin{tabular}{rrr|rrr|rrr}
			\hline
			Dataset          & image & mask & Dataset      & image & mask  & Dataset      & image & mask \\ \hline
			HE-BCT$^*$~\cite{Hemorrhage-BrainCTImages}           & 1,070                         & 316                          & LCTSC$^*$~\cite{LCTSC1}         & 14,338                        & 38,342      &	DZ Task07~\cite{DecathlonZ}        & 26,719                        & 11,320                 \\
			INSTANCE~\cite{INSTANCE1}         & 2,981                         & 892                          & COVID-19 Seg$^*$~\cite{COVID-19CTImageSeg}  & 929                          & 699   & CHAOS~\cite{CHAOS}        & 2,874                         & 2,341                       \\
			HaN-Seg~\cite{HaN-Seg}          & 7,581                         & 24,990                        & LNDB~\cite{LNDB1}         & 63,068                        & 2,369      & DZ Task09~\cite{DecathlonZ}        & 3,650                         & 1,050                       \\
			Totalsegmentator~\cite{Totalsegmentator} & 312,400                       & 3,837,723                      & ATM$^*$~\cite{ATM}           & 190,724                       & 124,231           & KiTS23~\cite{KiTS23}       & 95,221                        & 16,826                         \\
			HaN-OAR~\cite{HaN-OAR}          & 6,151                         & 13,356                        & Pe-CT-SEG~\cite{Pediatric-CT-SEG1}    & 110,083                       & 618,599     &	DZ Task10~\cite{DecathlonZ}        & 13,486                        & 1,280                            \\
			PDDCA$^*$~\cite{PDDCA}             & 7,367                         & 6,053                         & DZ Task08~\cite{DecathlonZ}    & 21,120                        & 12,852          & NIH-Pancreas~\cite{NIH-Pancreas1} & 18,942                        & 6,881                         \\
			COVID-19-20~\cite{COVID-19-20}      & 13,705                        & 4,956                         & AMOS22~\cite{AMOS22}       & 41,430                        & 129,237     & 	COVID-19 Scan$^*$~\cite{COVID-19CTscan}     & 3,520                         & 1,844                                \\
			MosMed~\cite{MosMed}           & 2,049                         & 785                          & WORD~\cite{WORD}        & 24,218                        & 59,192     & A-ACC-Ki67~\cite{Adrenal-ACC-Ki671}   & 6,008                         & 1,400                                      \\
			MM-WHS$^*$~\cite{MM-WHS}            & 5,305                         & 17,311                        & BTCV$^*$~\cite{BTCV}          & 3,779                         & 12,175    & FUMPE~\cite{FUMPE}            & 8,792                         & 2,304                               \\
			DZ Task06~\cite{DecathlonZ}        & 17,657                        & 1,646                         & LiTS~\cite{LiTS}         & 58,638                        & 26,269         & Pr-An-Ed-Ca$^*$~\cite{Prostate-Anatomical-Edge-Cases1}  & 23,359                        & 23,191                  \\ \hline      
		\end{tabular}
		}
		\label{tab2}
	\end{table*}
	
	The task-indicator prompt encoder takes task-specific positive and negative indicators, multi-resolution image embeddings from the CNN image encoder (\textit{i.e.}, $F_{16}$, $F_{32}$, $F_{64}$, $F_{128}$, and $F_{256}$), and the final image embedding $F_{vit}$ from the ViT image encoder as inputs to replace the manual prompt encoder in SAM and effortlessly generate positive point prompt embeddings, bounding box prompt embeddings, and negative point embeddings, as depicted in Fig.~\ref{fig3}. First, we pre-process the inputs to map their features to the feature space of interest for the indicator-specific task. Specifically, image embeddings are projected into the task-specific feature space via linear projections, forming a series of new image embeddings $\{F'\} = \{F'_{16}$, $F'_{32}$, $F'_{64}$, $F'_{128}$, $F'_{256}$, $F'_{vit}$\}. For a given task, a positive task-related indicator $P^+ \in \mathbb{R} ^ {1 \times d_p}$ and a negative task-related indicator $P^- \in \mathbb{R} ^ {1 \times d_p}$ are learned to generate the corresponding prompt embeddings. In each image embedding, as the maximum value usually reflects the most representative region for the given task (\textit{i.e.}, foreground) and the average value usually corresponds to background, we apply both max pooling and average pooling onto $\{F'\}$ and concatenate them to generate the positive image embedding $F^+$ and the negative image embedding $F^-$ respectively. Then, the negative point embedding ${Pe}^-$ is obtained by calculating cross-attention between $P^-$ and $F^-$, formulated as:
	\begin{equation}
		\begin{aligned}
			A^- = \text{\textbf{Softmax}}(\frac{P^-E_{q^-}(F^-E_{k^-})^T}{\sqrt{d_p}}), \\
			{Pe}^- = \text{\textbf{Gelu}}((A^-(F^-E_{v^-}))E_-E_{1})E_2,
		\end{aligned}
		\label{eq3}
	\end{equation}
	where $E_{q^-}, E_{k^-}, E_{v^-}, E_-, E_1$, and $E_2$ are projection matrices. Following Eq. \ref{eq3}, we can calculate $A^+$, the positive point embedding $Pe^+$, and the positive bounding box embedding $Be^+$. It is noted that not all inputs have foreground and providing positive prompts may be counter-productive. To address this, a classifier is added to the positive flow, formulated as
	\begin{equation}
		p = \text{\textbf{Softmax}}(A^+(F^+E_{v^+})E_+E_c),
		\label{eq4}
	\end{equation}
	where $E_c \in \mathbb{R} ^ {d_p \times 2}$ is a classification matrix. When $p$ indicates no foreground existing, the not-a-point embedding $Ne$ is used instead of the $Pe^+$ and $Be^+$ to prompt the mask decoder. $Ne$ is from the original SAM and frozen during training.
	
	\subsection{Training Strategies}
	
	Parameters of the shared parts between SAMCT and SAM inherit from SAM trained on SA-1B~\cite{sam}. During training SAMCT, only newly introduced parameters are updated through the standard Dice and binary cross-entropy losses. When training the task-indicator prompt encoder for any specific task, an additional cross-entropy loss is adopted to penalize $p$. When foreground exists, the prompts during training contain a randomly sampled positive point, a randomly sampled negative point, and a randomly shifted bounding box. Otherwise, the prompts only contain two randomly sampled negative points. For image-mask pairs converted from 3D data, as there exist many full background pairs, during most training rounds, we randomly select $10\%$ ($\leq$1000 pairs) of them for each object. The performance of SAMCT is evaluated under the prompts of a center positive point and a bounding box.
	
	\begin{table*}[!t]
		\centering
		\caption{Object Mapping. H: Head and Neck, C: Chest, A: Abdomen, P: Pelvis, B:Bone, L: lesion, V: vessel.}
		\resizebox{\textwidth}{!}{
		\begin{tabular}{ll|ll|ll|ll}
			\hline
			Object               & ID & Object           & ID & Object            & ID & Object            & ID \\ \hline
			brainstem                                & H1                          & left submandibular gland             & H2                          & right submandibular gland             & H3                          & optic chiasm                          & H4                          \\
			left optic nerve                         & H5                          & right optic nerve                    & H6                          & left parotid gland                    & H7                          & right parotid gland                   & H8                          \\
			spinal cord                              & H9                          & esophagus                            & H10                         & brain                                 & H11                         & buccal mucosa                         & H12                         \\
			oral cavity                              & H13                         & \makecell[l]{anterior segment\\ (left eyeball)} & H14                         & \makecell[l]{anterior segment\\ (right eyeball)} & H15                         & \makecell[l]{posterior segment\\ (left eyeball)} & H16                         \\
			\makecell[l]{posterior segment\\ (right eyeball)} & H17                         & left lacrimal gland                  & H18                         & right lacrimal gland                  & H19                         & lips                                  & H20                         \\
			pituitary gland                          & H21                         & face                                 & H22                         & left eye                              & H23                         & right eye                             & H24                         \\
			left lens                                & H25                         & right lens                           & H26                         & left temporal lobes                   & H27                         & right temporal lobes                  & H28                         \\
			left inner ear                           & H29                         & right inner ear                      & H30                         & left middle ear                       & H31                         & right middle ear                      & H32                         \\
			cricopharyngeal inlet                    & H33                         & thyroid                              & H34                         & larynx-glottis                        & H35                         & larynx-supraglottic                   & H36                         \\
			left ventricle                           & C1                          & right ventricle                      & C2                          & left atrium                           & C3                          & right atrium                          & C4                          \\
			myocardium                               & C5                          & right lung                           & C6                          & left lung                             & C7                          & trachea                               & C8                          \\
			body                                     & C9                          & left breast                          & C10                         & right breast                          & C11                         & thymus                                & C12                         \\
			pancreas                                 & A1                          & spleen                               & A2                          & right kidney                          & A3                          & left kidney                           & A4                          \\
			gallbladder                              & A5                          & liver                                & A6                          & stomach                               & A7                          & right adrenal gland                   & A8                          \\
			left adrenal gland                       & A9                          & bladder                              & A10                         & prostate                              & A11                         & rectum                                & A12                         \\
			duodenum                                 & A13                         & colon                                & A14                         & large Intestine                       & A15                         & large Intestine                       & A16                         \\
			gonads                                   & A17                         & skin                                 & A18                         & left autochthon                       & A19                         & right autochthon                      & A20                         \\
			utero cervix                             & P1                          & left gluteus maximus                 & P2                          & right gluteus maximus                 & P3                          & left gluteus medius                   & P4                          \\
			right gluteus medius                     & P5                          & left gluteus minimus                 & P6                          & right gluteus minimus                 & P7                          & left iliopsoas                        & P8                          \\
			right iliopsoas                          & P9                          & mandible                             & B1                          & left head of femur                    & B2                          & right head of femur                   & B3                          \\
			\makecell[l]{temporomandibular joint\\ (left)}             & B4                          & \makecell[l]{temporomandibular joint\\ (right)}        & B5                          & arytenoid                             & B6                          & left clavicula                        & B7                          \\
			right clavicula                          & B8                          & left humerus                         & B9                          & right humerus                         & B10                         & left rib                              & B11                         \\
			right rib                                & B12                         & left scapula                         & B13                         & right scapula                         & B14                         & vertebrae                             & B15                         \\
			spinal canal                             & B16                         & left hip                             & B17                         & right hip                             & B18                         & sacrum                                & B19                         \\
			hemorrhage                               & L1                          & COVID-19                             & L2                          & lung tumor                            & L3                          & pancreas tumor                        & L4                          \\
			colon cancer                             & L5                          & pulmonary embolism                   & L6                          & pulmonary nodule                      & L7                          & liver tumor                           & L8                          \\
			kidney tumor                             & L9                          & cyst                                 & L10                         & adrenocortical carcinoma              & L11                         & aorta                                 & V1                          \\
			pulmonary artery                         & V2                          & inferior vena cava                   & V3                          & portal and splenic vein          & V4                          & left carotid artery                   & V5                          \\
			right carotid artery                     & V6                          & hepatic vessel                       & V7                          & left iliac artery                     & V8                          & right iliac artery                    & V9                          \\
			left iliac vena                          & V10                         & right iliac vena                     & V11                         &                                       &                             &                                       &                             \\ \hline
		\end{tabular}
	}
		\label{tab3}
	\end{table*}	
	
	\section{Experiments}
	
	In this section, CT5M is first described in Sec. \ref{sec4.1}. The versatility and generalization of SAMCT trained on CT5M are evaluated on 30 datasets across 118 objects in Sec. \ref{sec4.2}. Then, SAMCT is compared against both foundation models in Sec. \ref{sec4.3} and SOTA specific models in Sec. \ref{sec4.4}. Component-wise ablation study of SAMCT are presented in Sec. \ref{sec4.5}. Finally, the effectiveness of the task-indicator prompt encoder is verified in Sec.  \ref{sec4.6}.
	
	\subsection {Dataset: CT5M}
	\label{sec4.1}
	
	CT5M is collected from 30 publicly available datasets as summarized in Table~\ref{tab2}. As both annotation protocols and data formats of datasets are different, a series of preprocessing operations are adopted to standardize the collected public datasets. For 2D data, all data is converted to the .PNG format for consistent data loading. For volumetric data, all 3D data is converted to 2D slices across the coronal plane to match the inputs of SAM-based models. As the intensity range of CT is usually (-1024, 2048), directly compressing such a large range to (0, 255) may lose valuable information. Therefore, under the guidance of experts, we choose different density windows for objects. Specifically, the intensity ranges of bone, tissue, lung, and hemorrhage are set as (-400, 1100), (-200, 300), (-1300, 300), and (-20, 100) respectively. 
	
	The collection of masks/objects has been carefully screened and checked to ensure that data is not duplicated and annotation rules are consistent. For instance, the adrenal glands are labeled as left and right adrenal glands in the AMOS dataset, while they are labeled as a whole in the WORD dataset. At this point, we only retain the fine left and right adrenal gland masks to remove annotation ambiguity. After the above preprocessing, CT5M collects 5M masks with foreground regions from 1.1M images of 30 publicly available datasets.
	
	According to the scanning position and object attribute, we divide objects in CT5M into seven categories, including head and neck (H), chest (C), abdomen (A), pelvis (P), lesion (L), bone (B), and vessel (V). Totally 118 objects from the above seven categories are included in CT5M. Details of objects and their mapping abbreviations are summarized in Table~\ref{tab3}.
	
	To conduct plentiful generalization and versatility verification, nine out of the preprocessed 30 datasets (\textit{i.e.}, datasets marked with $*$ in Table~\ref{tab2}) were set as training invisible, and the rest were selected as training visible. For training invisible datasets, all data is used as the testing set to evaluate the generalization ability of SAMCT. As for each training visible dataset, data is divided into training, validation, and testing sets. The segmentation performance of SAMCT on such visible testing sets reflects its versatility. Data partitioning of training visible datasets follows publicly available data splits. For datasets without public data partitioning or with only training set labels, they are randomly divided into training, validation, and testing sets in a 7:1:2 ratio. For datasets with only training and validation set labels, the validation set is divided into validation and testing sets in a 1:1 ratio.
	
	\subsection {Versatility and Generalization on CT5M}
	\label{sec4.2}
	
	\begin{figure}[!t]
		\centering{\includegraphics[width=1\linewidth]{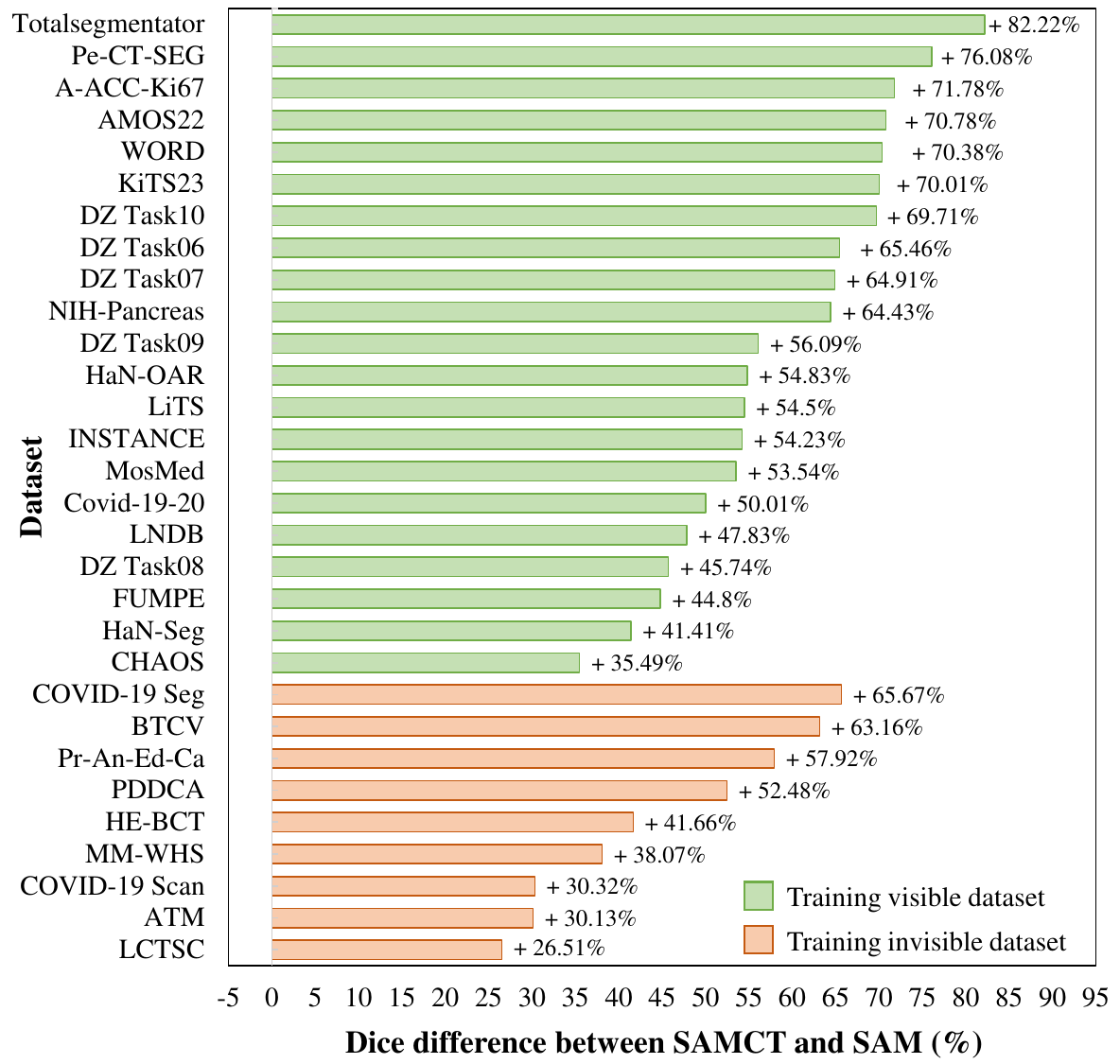}}
		\caption{
			Comparison between SAMCT and SAM across 21 training visible datasets (versatility) and 9 training invisible datasets (generalization). SAMCT consistently outperforms SAM with large margins.
		}
		\label{fig4}
	\end{figure}
	
	\begin{figure*}[!t]
		\centering{\includegraphics[width=\textwidth]{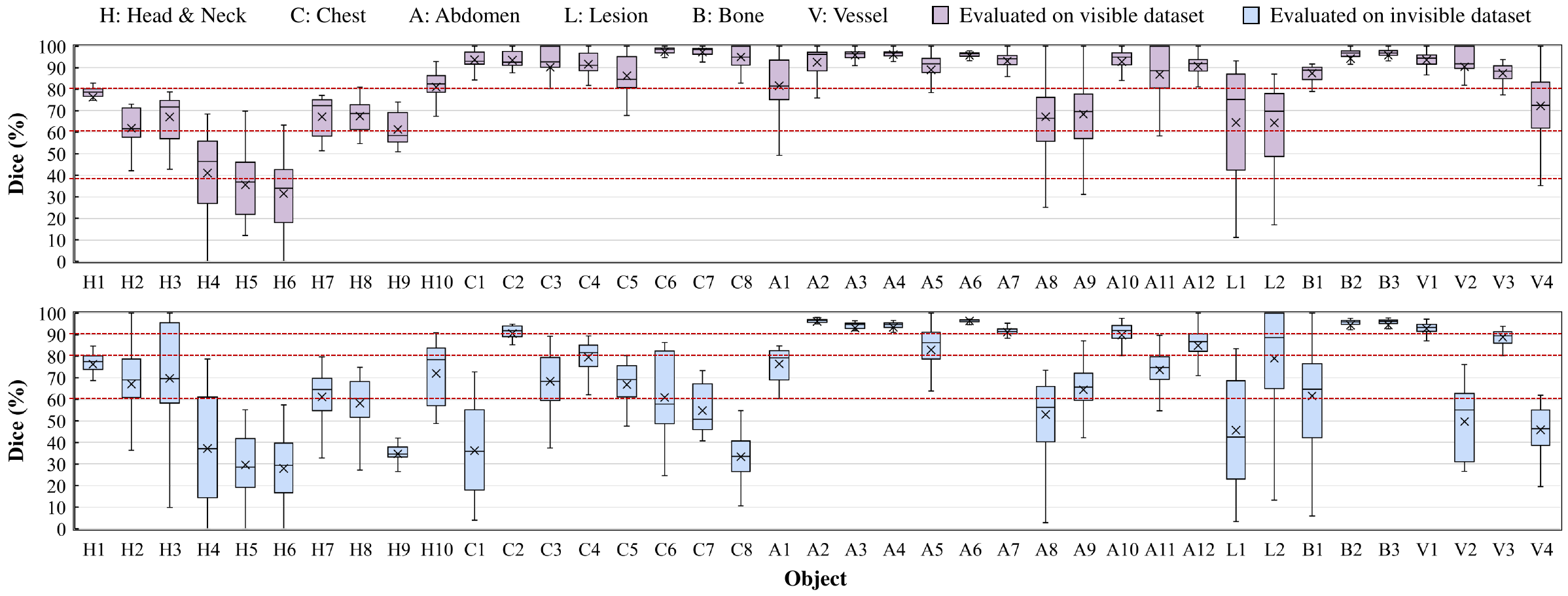}}
		\caption{
			Dice boxplot of SAMCT on visible and invisible datasets across 39 objects. Objects are represented by groups of category letters and ID numbers, and the object mapping table is provided in Table ~\ref{tab3}.
		}
		\label{fig5}
	\end{figure*}
	
	\begin{figure*}[!t]
		\centering{\includegraphics[width=\textwidth]{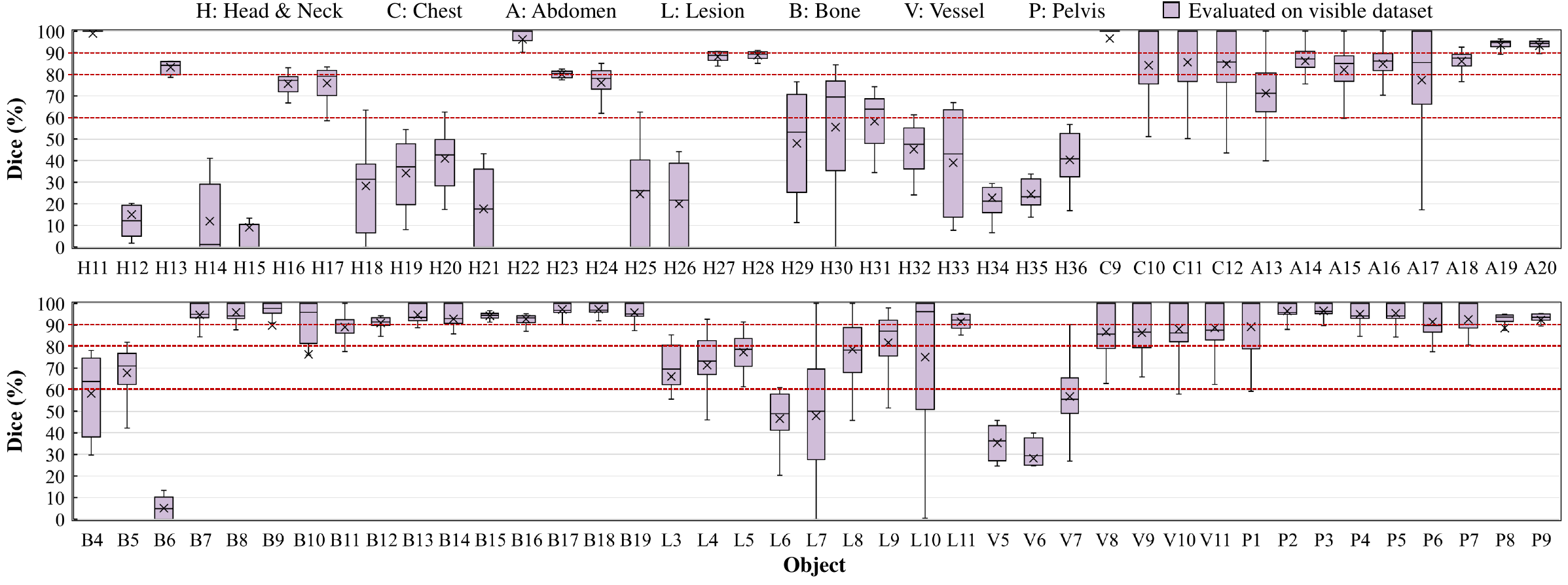}}
		\caption{
			Dice boxplot of SAMCT on visible datasets across the rest 79 objects. Objects are represented by groups of category letters and ID numbers, and the object mapping table is provided in Table ~\ref{tab3}.
		}
		\label{fig6}
	\end{figure*}
	
	\begin{figure}[!t]
		\centering{\includegraphics[width=\linewidth]{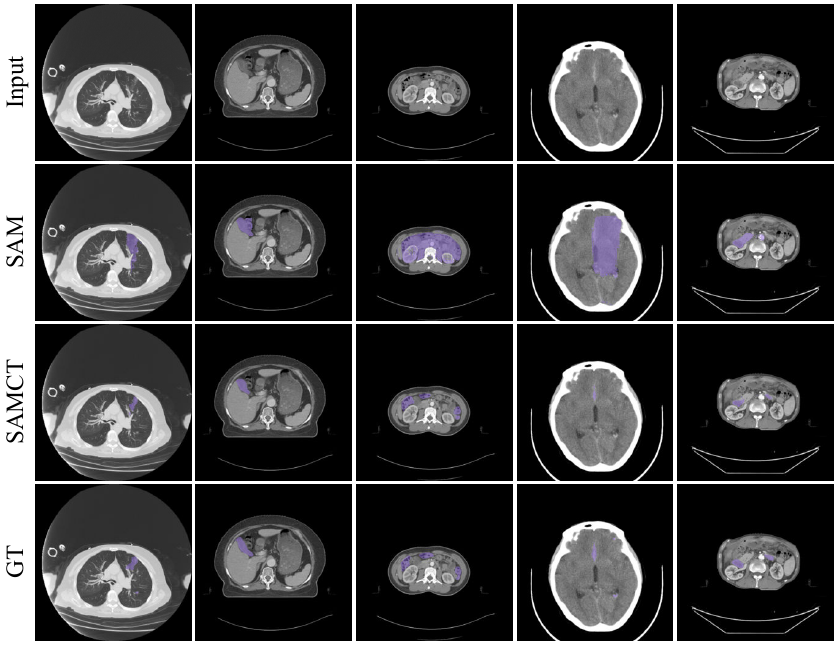}}
		\caption{
			Qualitative comparisons between SAM and SAMCT across 5 different objects. GT represents the ground truth.
		}
		\label{fig7}
	\end{figure}
	
	\noindent \textbf{Setting.} To train a powerful CT foundation model for auxiliary annotation and clinical applications, we trained SAMCT on the complete training data of CT5M and evaluated its versatility and generalization. During training, prompts were randomly sampled from positive points, negative points, and shifted bounding boxes generated from the masks. SAMCT was trained by an Adam optimizer with an initial learning rate of 0.0006 and a batch size of 48 for 50 epochs where the learning rate is adjusted to 0.0001 on the 10th epoch. Random rotation, random scaling, cropping, contrast adjustment, and gamma augmentation were applied for data augmentation.
	
	\begin{table*}[!t]
		\centering
		\caption{Dice comparison of foundation models on three representative training visible datasets, including COVID-19-20~\cite{COVID-19-20}, INSTANCE~\cite{INSTANCE1}, and WORD~\cite{WORD}, covering 16 objects. Note: ICH: intracranial hemorrhage, Liv: liver, Spl: spleen, LKid: left kidney, RKid: right kidney, Sto: stomach, Gall: gallbladder, Eso: esophagus, Pan: pancreas, Dou: duodenum, Col: colon, Rec: rectum, B: bladder, LHF: left head of femur, RHF: left head of femur, and Avg: the average Dice across 16 objects.}
		\resizebox{\textwidth}{!}{
			\begin{tabular}{l|cccccccccccccccc|c}
				\hline
				Method  & COVID-19       & ICH            & Liv            & Spl            & Lkid           & Rkid           & Sto            & Gal            & Eso            & Pan            & Duo            & Col            & Rec            & Bla            & LHF            & RHF            & Avg            \\ \hline
				SAM~\cite{sam}     & 15.55          & 10.26          & 41.88          & 22.55          & 20.00          & 19.62          & 22.76          & 2.02           & 1.66           & 6.87           & 5.69           & 16.66          & 10.39          & 17.85          & 24.06          & 23.93          & 16.36          \\
				SAM-Med2D~\cite{samed2d} & 69.69          & 73.11          & 95.24          & 94.16          & 94.12          & 94.26          & 92.14          & 78.14          & 78.88          & 80.94          & 73.15          & 79.76          & 89.27          & 94.27          & 94.35          & 94.33          & 85.99          \\
				MedSAM~\cite{medsam}  & 67.65          & 69.36          & 94.59          & 95.51          & 95.25          & 95.45          & 93.23          & \textbf{84.99} & 81.84          & 82.94          & 78.38          & 80.92          & 89.09          & 94.74          & 95.05          & 94.38          & 87.09          \\
				SAMed~\cite{samed}   & 77.33          & 74.36          & 95.24          & 93.68          & 93.52          & 92.82          & 92.92          & 83.18          & 82.07          & 84.63          & 79.57          & 83.34          & 90.43          & 94.07          & 94.03          & 93.42          & 87.79          \\
				MSA~\cite{msa}     & 76.54          & 77.58          & 96.04          & 95.33          & 95.19          & 95.23          & 93.57          & 83.75          & 83.74          & 84.59          & 79.72          & 85.35          & \textbf{91.54} & 94.89          & 95.29          & 95.42          & 88.99          \\
				\textbf{SAMCT}   & \textbf{77.59} & \textbf{77.66} & \textbf{96.35} & \textbf{95.86} & \textbf{95.61} & \textbf{95.82} & \textbf{93.99} & 83.87          & \textbf{85.03} & \textbf{86.02} & \textbf{82.14} & \textbf{86.29} & 91.29          & \textbf{95.33} & \textbf{95.85} & \textbf{96.09} & \textbf{89.67} \\ \hline
			\end{tabular}
		}
		\label{tab4}
	\end{table*}
	
	\noindent \textbf{Quantitative Results.} Comparison between SAMCT and SAM is depicted in Fig.~\ref{fig4}. Any bar in green has a positive value in Fig.~\ref{fig4} indicates that SAMCT outperforms SAM when tested on a training visible dataset, where CHAOS~\cite{CHAOS} is of the least performance improvements among training visible datasets, with an average increase of $35.49\%$ in Dice and Totalsegmentator~\cite{Totalsegmentator} is of the most performance improvements with an average increase of $82.22\%$ in Dice. It demonstrates the brilliant versatility of SAMCT on various downstream tasks. As for the generalization performance of SAMCT, all training invisible datasets own noticeable performance improvements, where LCTSC~\cite{LCTSC1} is of the least improvements with an average increase of $26.51\%$ in Dice and COVID-19 Seg~\cite{COVID-19CTImageSeg} is of the most performance improvements with an average increase of $65.67\%$ in Dice. It proves the prominent generalization ability of SAMCT. 
	
	To analyze the object-specific performance of SAMCT, we calculate and plot the average Dice score of each object in Figs.~\ref{fig5} and \ref{fig6}. For training visible datasets, most objects in chest, abdomen, bone, and pelvis have an average Dice score greater than $90\%$. Specifically, 44 out of 118 objects have an average Dice score greater than 90$\%$ and 68 out of 118 objects have an average Dice score greater than $80\%$, validating the surprising versatility of SAMCT. Among 4,236,340 training image-mask pairs, the numbers of masks for H4, H5, H6, H12, H14, and H15 are only 146, 235, 241, 448, 186, and 201, respectively, resulting in relatively poor performance of SAMCT. Similarly, B5 and some other objects in head and neck exhibit poorer performance due to relatively limited training data. For training invisible datasets, 16 and 10 out of 39 objects have an average Dice score greater than $80\%$ and $90\%$, making SAMCT a powerful tool for clinically-assisted annotation. 
	
	\noindent \textbf{Qualitative Results.} Qualitative segmentation results of SAM and SAMCT are illustrated in Fig.~\ref{fig7}. Visually, segmenting various medical objects via one model is challenging, especially for those with low contrast, blurry boundaries, small sizes, and complex shapes. Compared to SAM, SAMCT can accurately identify such hard objects, validating the effectiveness of constructing the large CT dataset (\textit{i.e.}, CT5M), the U-shaped CNN image encoder, and the cross-branch interaction in SAMCT.
	
	\subsection {Comparison with Foundation Models}
	\label{sec4.3}
	
	\noindent \textbf{Setting.} Experiments are conducted on a subset of CT5M, consisting of COVID-19-20~\cite{COVID-19-20}, INSTANCE~\cite{INSTANCE1}, and WORD~\cite{WORD} and covering 16 objects. For a fair comparison, all models are initialized with SAM and prompted by the joint of random point and random box prompts in training. During inference, all models generate masks under the prompt of the joint center point and bounding box. All models were trained by an Adam optimizer with an initial learning rate of $1e^{-4}$ and a batch size of 8. A separate BTCV~\cite{BTCV} dataset with overlapping objects of WORD, \textit{i.e.}, liver, spleen, left kidney, right kidney, stomach, gallbladder, esophagus, and pancreas, are used for generalization verification. 
	
	\noindent \textbf{Results.} Quantitative comparison results of different foundation models on three representative training visible datasets are summarized in Table~\ref{tab4}. Among SAM-based models for medicine, all models can greatly improve the performance of SAM by learning more medical knowledge, where MedSAM~\cite{medsam} and MSA~\cite{msa} achieve the best performance in gallbladder and rectum segmentation, respectively. Besides, MSA achieves the best average Dice across 16 objects among comparison models. Comparatively, SAMCT achieves the second-best performance in the segmentation of gallbladder and rectum and outperforms all comparison models on the other 14 objects, leading to the best average Dice score of $89.67\%$. Quantitative comparison results of different foundation models on the training invisible BTCV~\cite{BTCV} dataset are summarized in Table~\ref{tab5}. SAM-based medical models are still much superior compared to SAM on generalization, especially for small-size objects, \textit{e.g.}, gallbladder and esophagus, and low-contrast objects, \textit{e.g.}, stomach. Specifically, MSA and MedSAM achieve the best performance in segmenting esophagus and right kidney, with the Dice scores of $84.15\%$ and $90.69\%$, respectively. Compared to MSA and MedSAM, though SAMCT performs slightly worse on esophagus and right kidney, it is superior on the other 6 objects, leading to an improvement of $5.06\%$ and $1.75\%$ in average Dice across 8 objects. It should be noted that the best comparison models for training visible and invisible validation are different (\textit{i.e.}, MSA and MedSAM respectively) while SAMCT consistently achieves the best overall performance, demonstrating the brilliant versatility and generalization ability of SAMCT.
	
	\subsection {Comparison with Task-Specific Models}
	\label{sec4.4}
	\noindent \textbf{Setting.} Comparison models include SAMCT trained on a subset of CT5M (denoted as SAMCT-Sub obtained in Sec.~\ref{sec4.3}), SAMCT trained on CT5M (denoted as SAMCT-CT5M obtained in Sec.~\ref{sec4.2}), and 13 SOTA task-specific models. Task-specific models consist of four 2D CNN-based methods, \textit{i.e.}, U-Net~\cite{unet}, CPFNet~\cite{cpfnet}, CA-Net~\cite{canet}, and AAU-net~\cite{aaunet}, four 2D transformer-based methods, \textit{i.e.}, TransFuse~\cite{transfuse}, TransUNet~\cite{transunet}, MISSFormer~\cite{missformer}, and H2Former~\cite{h2former}, and five 3D methods, \textit{i.e.}, nnU-Net~\cite{nnunet}, SwinUNETR~\cite{swinunetr}, UNETR~\cite{unetr}, nnFormer~\cite{nnformer}, and MedNeXt~\cite{mednext}. All task-specific models were trained on INSTANCE~\cite{INSTANCE1} and WORD~\cite{WORD} respectively following the same settings with SAMCT.
	
	\begin{table*}[!t]
		\centering
		\caption{Dice comparison of foundation models on the training invisible BTCV~\cite{BTCV} dataset.}
		\setlength{\tabcolsep}{10pt}
		\resizebox{\textwidth}{!}{
			\begin{tabular}{l|cccccccc|c}
				\hline
				Method  & Liv            & Spl            & Lkid           & Rkid           & Sto            & Gal            & Eso            & Pan            & Avg   \\ \hline
				SAM~\cite{sam}     & 56.51          & 30.05          & 27.34          & 27.96          & 28.34          & 4.01           & 2.54           & 32.99          & 26.22 \\
				SAM-Med2D~\cite{samed2d} & 83.54          & 91.60          & 90.10          & 90.22          & 75.57          & 76.47          & 82.74          & 57.39          & 80.95 \\
				MSA~\cite{msa}     & 81.90          & 92.79          & 89.74          & 89.58          & 80.39          & 70.77          & \textbf{84.15} & 60.75          & 81.26 \\
				SAMed~\cite{samed}   & 81.73          & 91.85          & 87.37          & 88.33          & 80.57          & 77.67          & 80.57          & 66.42          & 81.81 \\
				MedSAM~\cite{medsam}  & 88.33          & 90.44          & 90.52          & \textbf{90.69} & 81.29          & 83.48          & 84.06          & 67.73          & 84.57 \\
				\textbf{SAMCT}   & \textbf{89.98} & \textbf{95.29} & \textbf{90.98} & 90.56          & \textbf{85.23} & \textbf{84.15} & 83.74          & \textbf{70.63} & \textbf{86.32} \\ \hline
			\end{tabular}
		}
		\label{tab5}
	\end{table*}
	
	\begin{table}[!t]
		\centering
		\caption{Comparison of SAMCT and SOTA task-specific models on the training visible INSTANCE~\cite{INSTANCE1} and WORD~\cite{WORD} datasets. }
		\resizebox{\linewidth}{!}{
			\begin{tabular}{l|ccc|ccc}
				\hline
				\multirow{2}{*}{Method} & \multicolumn{3}{c|}{INSTANCE}                   & \multicolumn{3}{c}{WORD}                        \\ \cline{2-7}
				& Dice           & HD            & IoU            & Dice           & HD            & IoU            \\ \hline
				U-Net~\cite{unet}                   & 51.91          & 22.75         & 44.03          & 86.53          & 17.04         & 78.47          \\
				CPFNet~\cite{cpfnet}                  & 52.30           & 22.45         & 43.58          & 83.74          & 18.41         & 74.64          \\
				CA-Net~\cite{canet}                  & 55.07          & 22.30          & 45.29          & 85.86          & 17.35         & 77.59          \\
				AAU-net~\cite{aaunet}                 & 56.91          & 20.30          & 48.24          & 86.63          & 16.46         & 78.42          \\ \hline
				TransFuse~\cite{aaunet}               & 44.42          & 24.40          & 35.54          & 74.39          & 20.87         & 62.44          \\
				TransUnet~\cite{transunet}               & 51.75          & 22.10          & 43.42          & 85.38          & 17.37         & 76.84          \\
				MISSFormer~\cite{missformer}              & 58.18          & 21.35         & 48.26          & 84.16          & 17.99         & 75.16          \\
				H2Former~\cite{h2former}                & 58.22          & 19.70          & 48.83          & 85.79          & 17.82         & 77.31          \\ \hline
				SwinUNETR~\cite{swinunetr} & 60.00 &18.85 &49.99 &73.23 &14.29 &62.66 \\
				nnFormer~\cite{nnformer} & 60.61 &18.45 &50.76 &78.94 &13.73 &67.55 \\
				UNETR~\cite{unetr} & 60.97 &18.75 &50.81 &75.27 &13.63 &65.38 \\
				nnU-Net~\cite{nnunet} & 63.18 &18.45 &52.91 &84.52 &12.09 &75.27  \\
				MedNeXt~\cite{mednext} & 64.14 &18.05 &53.80 &83.73 &12.77 &74.06 \\
				\hline
				\textbf{SAMCT-Sub}                 & \textbf{77.66} & \textbf{15.90} & \textbf{65.76} & \textbf{91.40} & 12.29         & 84.73          \\
				\textbf{SAMCT-CT5M}              & 64.49          & 19.45         & 52.88          & 87.23          & \textbf{4.33} & \textbf{87.57} \\ \hline
			\end{tabular}
		}
		\label{tab6}
	\end{table}
	
	\begin{table*}[!t]
		\centering
		\caption{Dice comparison of SAMCT and SOTA specific models on the training invisible BTCV~\cite{BTCV} dataset.}
		\setlength{\tabcolsep}{10pt}
		\resizebox{\textwidth}{!}{
		\begin{tabular}{l|cccccccc|c}
			\hline
			Method     & Liv            & Spl            & Lkid           & Rkid           & Sto            & Gal            & Eso            & Pan            & Avg            \\ \hline
			CPFNet~\cite{cpfnet}     & 64.25          & 4.96           & 3.13           & 1.34           & 24.75          & 0.00           & 6.24           & 0.00           & 13.08          \\
			U-Net~\cite{unet}      & 77.22          & 19.46          & 22.80          & 15.80          & 12.89          & 3.52           & 7.92           & 1.50           & 20.14          \\
			CA-Net~\cite{canet}     & 70.77          & 28.18          & 10.82          & 16.04          & 23.46          & 0.54           & 31.02          & 3.42           & 23.03          \\
			AAU-net~\cite{aaunet}    & 78.18          & 30.96          & 29.60          & 32.18          & 19.61          & 0.14           & 25.29          & 1.50           & 27.18          \\ \hline
			TransUNet~\cite{transunet}  & 68.67          & 9.46           & 1.71           & 0.74           & 19.08          & 3.58           & 2.43           & 1.60           & 13.41          \\
			TransFuse~\cite{transfuse}  & 77.13          & 4.04           & 4.73           & 3.76           & 14.55          & 3.33           & 0.40           & 0.01           & 13.49          \\
			MISSFormer~\cite{missformer} & 65.45          & 21.74          & 8.75           & 2.99           & 15.35          & 3.68           & 9.26           & 2.68           & 16.24          \\
			H2Former~\cite{h2former}   & 77.34          & 21.57          & 0.84           & 2.47           & 36.95          & 0.15           & 13.28          & 4.32           & 19.62          \\ 
			\hline
			SwinUNETR~\cite{swinunetr} & 84.24 &68.03 &69.36 &74.58 &54.11 &32.29 &38.19 &30.44 &56.41 \\
			nnFormer~\cite{nnformer} & 85.16&72.20 &69.89 &75.76 &54.60 &44.08 &44.04 &33.59 &59.91\\
			UNETR~\cite{unetr} & 84.49 &66.32 &71.66 &76.83 &55.22 &30.32 &36.60 &32.98 &56.80\\
			nnU-Net~\cite{nnunet} & 91.03 &78.55 &84.93 &85.87 &67.63 &59.89 &75.09 &54.09 &74.64\\
			MedNeXt~\cite{mednext} & 90.09 &82.35 &79.26 &80.34 &63.58 &55.91 &67.96 &47.73 &70.90 \\
			\hline
			\textbf{SAMCT-Sub}    & 89.98          & 95.29          & 90.98          & 90.56          & 85.23          & \textbf{84.15} & \textbf{83.74} & 70.63          & 86.32          \\
			\textbf{SAMCT-CT5M} & \textbf{96.33} & \textbf{96.13} & \textbf{93.38} & \textbf{93.13} & \textbf{91.14} & 82.99          & 82.72          & \textbf{91.42} & \textbf{90.91} \\ \hline
		\end{tabular}
	}
		\label{tab7}
	\end{table*}
	
	\noindent \textbf{Results.} Quantitative comparison results of different methods on visible INSTANCE~\cite{INSTANCE1} and WORD~\cite{WORD} are summarized in Table~\ref{tab6}. Among task-specific methods, MedNeXt~\cite{mednext} achieves the best performance on INSTANCE and nnU-Net~\cite{nnunet} achieves the best performance on WORD. Comparatively, SAMCT trained on either partial or complete CT5M achieves better performance than all task-specific methods across the two datasets. Quantitative comparison results of different methods on the invisible BTCV~\cite{BTCV} dataset are summarized in Table~\ref{tab7}. Obviously, task-specific methods perform poorly on invisible data due to domain shift. Comparatively, the performance of SAMCT is much better, with average Dice scores of $86.32\%$ and $90.91\%$ by SAMCT-Sub and SAMCT-CT5M, respectively. Benefiting from the construction of CT5M, though SAMCT-CT5M performs slightly worse in visible datasets due to the lower proportion of training targets compared to SAMCT-Sub, its performance on invisible datasets is much greater than that of SAMCT-Sub.
	
	\begin{table}[!t]
		\centering
		\caption{Component-wise ablation study of SAMCT on the COVID-19-20~\cite{COVID-19-20} dataset. CNN, CBI, and Adapter represent the U-shaped CNN image encoder, cross-branch interaction, and adapters in the ViT image encoder.}
		\begin{tabular}{cccccc}
			\hline
			CNN & CBI & Adapter & Dice           & HD             & IOU            \\ \hline
			$\bullet$   & $\bullet$   & $\bullet$       & \textbf{76.20} & 27.56          & \textbf{62.43} \\
			$\bullet$   & $\bullet$   & $\circ$       & 75.35          & 28.28          & 61.39          \\
			$\bullet$   & $\circ$   & $\bullet$       & 75.63          & \textbf{26.03} & 61.62          \\
			$\circ$   & $\circ$   & $\bullet$       & 74.90          & 31.23          & 61.21          \\ \hline
		\end{tabular}
		\label{tab8}
	\end{table}
	
	\subsection{Component-Wise Ablation Study}
	\label{sec4.5}
	
	Component-wise ablation study of SAMCT is summarized in Table~\ref{tab8}. Generally, removing any component is harmful, validating the effective designs of SAMCT.
	
	\subsection {Comparison with Various Prompt Modes}
	\label{sec4.6}
	
	\noindent \textbf{Setting.} We conduct a comparison between task-indicator prompt and various manual prompts, including the randomly sampled point prompt, the center point prompt, the random box prompt with an offset of less than $5\%$ width/height, the bounding box prompt, and the joint center point and bounding box prompt. SAMCT-CT5M is evaluated on WORD~\cite{WORD} under the above prompts. The task-indicator prompt encoder is a plug-in module independent of SAMCT and trained on the frozen SAMCT-CT5M with an initial learning rate of 0.0003 and a batch size of 12.
	
	\noindent \textbf{Results.} Quantitative comparison under various prompts is summarized in Table~\ref{tab9}. In terms of the random point prompt and the center point prompt, the former achieves better results in segmenting liver, spleen, and left kidney while the latter is better on segmenting right kidney, left head of femur, and right head of femur. In other words, the position of point prompts has a crucial impact on the performance of SAM-based methods, and no point type has been proven to be the best for all. In contrast, the segmentation results generated under different box prompts are relatively stable and better than various point prompts. Besides, using the joint center point and bounding box prompt consistently achieves the best performance among manual prompt modes. Comparatively, the proposed task-indicator prompt consistently outperforms any point prompt, achieving better performance than the random box prompt in segmenting liver, left kidney, right kidney, and left head of femur, approaching the best manual prompt in segmenting liver, spleen, left head of femur, and right head of femur, and even outperforming the best manual prompt in segmenting left kidney and right kidney. Though the task-indicator prompt cannot completely surpass the best manual prompt, it is labor-free and more easy-to-use in clinical applications. In addition, as a plug-and-play module, it is very lightweight with only 1.7M parameters, being highly extendable to other foundation models.
	
	{\renewcommand{\arraystretch}{1.0}
		\begin{table}[!t]
			\centering
			\caption{Dice comparison of different prompts on objects from the WORD dataset. Cpoint+bbox represents using the joint center point and bounding box prompt.}
			\resizebox{1\linewidth}{!}{
				\begin{tabular}{l|cccccc}
					\hline
					Prompt Mode      & Liv            & Spl            & Lkid           & Rkid          & LHF  & RHF         \\ \hline 
					random ponit         & 90.18          & 93.82          & 93.50           & 91.40          & 90.11    & 91.54      \\
					center point         & 86.71          & 93.6           & 92.77          & 93.60          & 90.69   & 91.94       \\ \hline
					random box           & 95.27          & 94.57          & 93.62          & 93.66         & 92.90  & 93.66       \\
					bounding box         & 95.46          & 94.67          & 93.77          & 93.90          & 93.05 & 93.72         \\ \hline
					cpoint+bbox          & \textbf{95.49} & \textbf{94.82} & 94.23          & 94.22         & \textbf{93.77}  & \textbf{93.98}\\ \hline
					\textbf{task-indicator} & 95.30           & 94.03          & \textbf{94.29} & \textbf{94.50} & 92.96   & 92.99       \\ \hline
				\end{tabular}
			}
			\label{tab9}
		\end{table}
	}
	
	\section{Conclusion}
	
	In this work, we present a large CT image dataset and a novel segmentation foundation model named SAMCT for various CT-related segmentation tasks. Different from existing research on tuning SAM on medical data, SAMCT has three unique components, including a U-shaped CNN image encoder for local feature supplement, a cross-branch interaction module for knowledge transfer, and a task-indicator prompt encoder for labor-free interaction. It deserves mentioning that the task-indicator prompt provides a new paradigm for labor-free prompts, converting semi-automatic interaction in SAM into fully-automatic interaction in SAMCT, which is more convenient and favorable in clinical applications. Extensive comparison experiments demonstrate the extraordinary versatility and generalization ability of SAMCT, outperforming SOTA both medical foundation models and task-specific models on various tasks, even without manual prompts.
	
	\bibliography{samctbib}
	
\end{document}